\documentclass[conference]{IEEEtran}
\IEEEoverridecommandlockouts
\usepackage{cite}
\usepackage{amsmath,amssymb,amsfonts}
\usepackage{algorithmic}
\usepackage{graphicx}
\usepackage{textcomp}
\usepackage{multirow}
\usepackage{booktabs}
\usepackage{url} 
\usepackage{xcolor}
\usepackage[hidelinks]{hyperref}
\def\BibTeX{{\rm B\kern-.05em{\sc i\kern-.025em b}\kern-.08em
    T\kern-.1667em\lower.7ex\hbox{E}\kern-.125emX}}
\begin{document}

\title{Volumetric Directional Diffusion: Anchoring Uncertainty Quantification in Anatomical Consensus for Ambiguous Medical Image Segmentation}

\author{
\IEEEauthorblockN{
Chao Wu, Mahesh Bhosale, Kangxian Xie, Pouya Karimian, David Doermann, Mingchen Gao
}
\IEEEauthorblockA{
\textit{Department of Computer Science and Engineering}\\
\textit{University at Buffalo, SUNY}\\
Buffalo, NY, USA\\
\{chaowu, mbhosale, kangxian, pouyakar, doermann, mgao8\}@buffalo.edu
}
}

\maketitle

\begin{abstract}
Ambiguous 3D medical image segmentation often involves boundaries where different expert delineations are non-identical yet clinically plausible. Modeling such inter-observer variability requires a careful balance between diversity and anatomical fidelity: deterministic models preserve coherent volumetric structures but collapse expert disagreement into a single mask, while stochastic generative models can produce diverse samples but may introduce disconnected components or slice-to-slice inconsistency when generating full 3D masks from unstructured noise. We propose Volumetric Directional Diffusion (VDD), a prior-anchored diffusion framework that shifts stochastic generation from full-mask synthesis to residual boundary exploration. VDD uses a coarse consensus prediction as an anatomical anchor and learns a directional diffusion process to generate plausible boundary variations around ambiguous regions while preserving stable volumetric topology. Experiments on three multi-rater datasets, including LIDC-IDRI, KiTS21, and ISBI 2015, show that VDD improves uncertainty distribution alignment while maintaining competitive segmentation accuracy and 3D structural consistency. These results suggest that prior-anchored residual diffusion can model clinically plausible expert disagreement without sacrificing anatomical fidelity.
\end{abstract}


\section{Introduction}

Accurate boundary delineation in volumetric medical imaging is notoriously challenging. 
For equivocal structures (e.g., ground-glass opacities, infiltrative boundaries), different experts may provide non-identical yet clinically plausible contours, reflecting genuine inter-observer variability rather than simple annotation noise. 
Capturing such aleatoric uncertainty is critical for downstream tasks such as radiotherapy planning and surgical margin assessment \cite{sha2025}, where boundary misestimation may carry severe clinical risks. 

However, this objective creates a core trade-off for segmentation models: they must represent the diversity of clinically plausible expert delineations while preserving coherent 3D anatomical structures. 
State-of-the-art deterministic 3D pipelines (e.g., nnU-Net \cite{isensee2021nnunet}) typically excel at producing spatially consistent masks, but they achieve this by collapsing multiple plausible annotations into a single consensus prediction, thereby suppressing inter-rater variability and yielding over-confident boundaries.

To model inter-reader variability, generative models were introduced, pioneering with VAE-based approaches such as the Probabilistic U-Net \cite{kohl2018probabilistic}. 
Recently, Denoising Diffusion Probabilistic Models (DDPMs)~\cite{ho2020denoising} have gained significant traction as general medical segmentation solvers \cite{wu2022medsegdiff,Lyu2025Step}. 
By formulating segmentation as a conditional generation task, diffusion models can produce a diverse set of segmentation masks by sampling from a learned distribution \cite{amit2021segdiff,wolleb2022diffusion,wu2025model,jin2025semantic}. 
This stochastic capability has been successfully exploited to model aleatoric uncertainty from divergent annotations \cite{rahman2023ambiguous,zbinden2023stochastic} and to establish multi-expert consensus \cite{zhang2024diffoseg}. 
Diffusion processes have also demonstrated potential in refining noisy or incomplete 3D clinical labels \cite{fan2025noisy,zhang2024molfescue}.
% zhang
Despite their flexibility, directly applying standard diffusion formulations to ambiguous 3D segmentation may be inefficient for the specific goal of modeling boundary uncertainty. 
In a conventional diffusion process, the model recovers the segmentation mask from an isotropic, independently distributed Gaussian state.
For high-dimensional 3D voxel masks, this formulation provides a broad generative search space in which both the global anatomical support and local boundary details must be reconstructed \cite{wdm2024, wusimplex}.

This setting is not fully aligned with ambiguous volumetric segmentation, where the coarse object location and macroscopic topology are often relatively stable, while clinically meaningful uncertainty is concentrated around equivocal boundaries. 
Without an explicit volumetric reference, stochastic generation may allocate capacity to unnecessary global shape variation rather than focusing on fine-grained boundary disagreement, making slice-to-slice consistency harder to maintain \cite{lyu2024slice,wang2025emssd}. 
This motivates a more constrained formulation that preserves coarse anatomical support while modeling residual boundary variations.

Motivated by this observation, we propose \textbf{Volumetric Directional Diffusion (VDD)}, which shifts the generative paradigm from full-mask synthesis to residual boundary exploration. 
Rather than treating a deterministic segmentation as a perfect pseudo-ground truth, VDD uses it as a coarse anatomical reference that provides macroscopic spatial support while allowing stochastic refinement around ambiguous boundaries. 
This design focuses generative stochasticity on regions where experts may reasonably disagree, while preserving stable volumetric topology for anatomically coherent 3D segmentation. 
Our main contributions are threefold:

\begin{itemize}
    \item {First}, we formulate ambiguous 3D medical image segmentation as a problem of balancing clinically plausible inter-rater variability with volumetric anatomical fidelity. 
    This perspective motivates residual boundary exploration as a more constrained and clinically meaningful alternative to generating full 3D masks from unstructured Gaussian noise.
    
    \item {Second}, we introduce \textit{Anatomical Anchoring}, which mathematically reformulates the diffusion trajectory \cite{hou2025directional} to incorporate a deterministic structural prior as a coarse anatomical reference. 
    By restricting the generative search space to \emph{residual exploration}, VDD aims to reduce topological fractures and promote slice-to-slice volumetric consistency.
    
    \item {Third}, extensive experiments on three multi-rater datasets (LIDC-IDRI, KiTS21, ISBI 2015) demonstrate that VDD achieves strong uncertainty modeling performance, as measured by metrics such as GED and CI, while preserving competitive segmentation accuracy and macroscopic structural consistency.
\end{itemize}

\section{Related Work}
\subsection{Ambiguous Segmentation and Multi-rater Labels}

Medical image segmentation can be inherently ambiguous when anatomical boundaries are low-contrast, infiltrative, or affected by pathology-specific variations. 
In such cases, multiple expert annotations may be clinically plausible rather than simply noisy, motivating models that represent segmentation as a distribution of possible masks instead of a single deterministic target. 
Probabilistic U-Net is an early representative approach that combines a U-Net backbone with a conditional variational latent space, enabling multiple plausible segmentation hypotheses to be sampled from the same input image~\cite{kohl2018probabilistic}. 
PHiSeg extends this idea with a hierarchical probabilistic formulation, where latent variables at different resolutions model segmentation uncertainty across multiple spatial scales~\cite{baumgartner2019phiseg}. 
Stochastic Segmentation Networks (SSN) further model spatially correlated aleatoric uncertainty through a low-rank multivariate Gaussian distribution in logit space, allowing structured hypotheses to be generated beyond independent pixel-wise uncertainty estimates~\cite{monteiro2020stochastic}. 
These probabilistic and stochastic approaches establish that inter-rater variability should be modeled rather than collapsed into a single annotation. 
Our work builds on this ambiguity-aware perspective, but focuses on volumetric medical segmentation, where sampled hypotheses should reflect plausible expert disagreement while preserving coherent 3D anatomical support.

\subsection{Diffusion-based Ambiguous Segmentation}

Diffusion models have recently been widely explored for medical image segmentation, with representative designs including iterative mask generation, dynamic conditional encoding, Bernoulli diffusion, hybrid diffusion refinement, and autoregressive mask prediction~\cite{amit2021segdiff,wu2022medsegdiff,chen2023berdiff,chen2024hidiff,chen2025autoregressive}. 
While these studies demonstrate the flexibility of diffusion-based generative modeling for segmentation, not all of them are designed for ambiguous or multi-rater settings. 
Here we focus on diffusion-based methods that explicitly model stochastic segmentation or expert disagreement.

Rahman et al. propose a diffusion-based framework for ambiguous medical image segmentation, generating multiple plausible masks to capture the distribution of expert annotations~\cite{rahman2023ambiguous}. 
CCDM formulates stochastic segmentation as conditional categorical diffusion, enabling the generation of multiple label maps that account for aleatoric uncertainty in divergent ground-truth annotations~\cite{zbinden2023stochastic}. 
DiffOSeg further studies ambiguous medical segmentation under a multi-expert setting, using diffusion to model consensus and expert-specific preferences~\cite{zhang2024diffoseg}. 
These methods are closely related to our goal of modeling plausible segmentation variability rather than producing a single deterministic mask.

Despite this progress, ambiguous 3D segmentation introduces an additional challenge: clinically meaningful disagreement is often concentrated around equivocal boundaries, while the coarse anatomical support of the target structure may remain relatively stable. 
Existing diffusion-based ambiguous segmentation methods generally introduce stochasticity at the level of the full mask or label map. 
For volumetric segmentation, this can make the generative search space broader than necessary when the main uncertainty lies in boundary deviations rather than global object presence or topology. 
VDD addresses this setting by shifting the diffusion variable from the full mask to the residual field around a case-specific anatomical reference, so that stochastic sampling focuses on plausible boundary variation while retaining macroscopic volumetric support.

\section{Methodology}

\subsection{Problem Setup and Anatomical Reference}

Let $x \in \mathbb{R}^{C \times D \times H \times W}$ denote a 3D medical image, and let 
$\mathcal{Y}=\{y^{(k)}\}_{k=1}^{K}$ denote the set of expert annotations for the same case. 
Each annotation $y^{(k)} \in \{0,1\}^{D \times H \times W}$ represents one clinically plausible delineation. 
During training, we randomly sample one annotation $y_0$ from $\mathcal{Y}$ as the diffusion target.

VDD first obtains a coarse anatomical reference $\hat{y}$ from a deterministic segmentation network $f_\phi$:
\begin{equation}
    \hat{y}=f_\phi(x).
\end{equation}
The network $f_\phi$ is trained on a consensus annotation derived from the available expert masks. 
We emphasize that $\hat{y}$ is not treated as a perfect pseudo-ground truth. 
Instead, it provides a case-specific estimate of the macroscopic object support, while boundary-level deviations from this reference are modeled by the stochastic diffusion process.
In the following, $\hat{y}$ denotes the deterministic anatomical reference,
$y_t$ denotes an intermediate noisy mask state at timestep $t$, and
$\tilde{y}^{(n)}$ denotes the final mask sampled from the $n$-th reverse trajectory.

\begin{figure*}[t]
\centering
\includegraphics[width=0.91\textwidth]{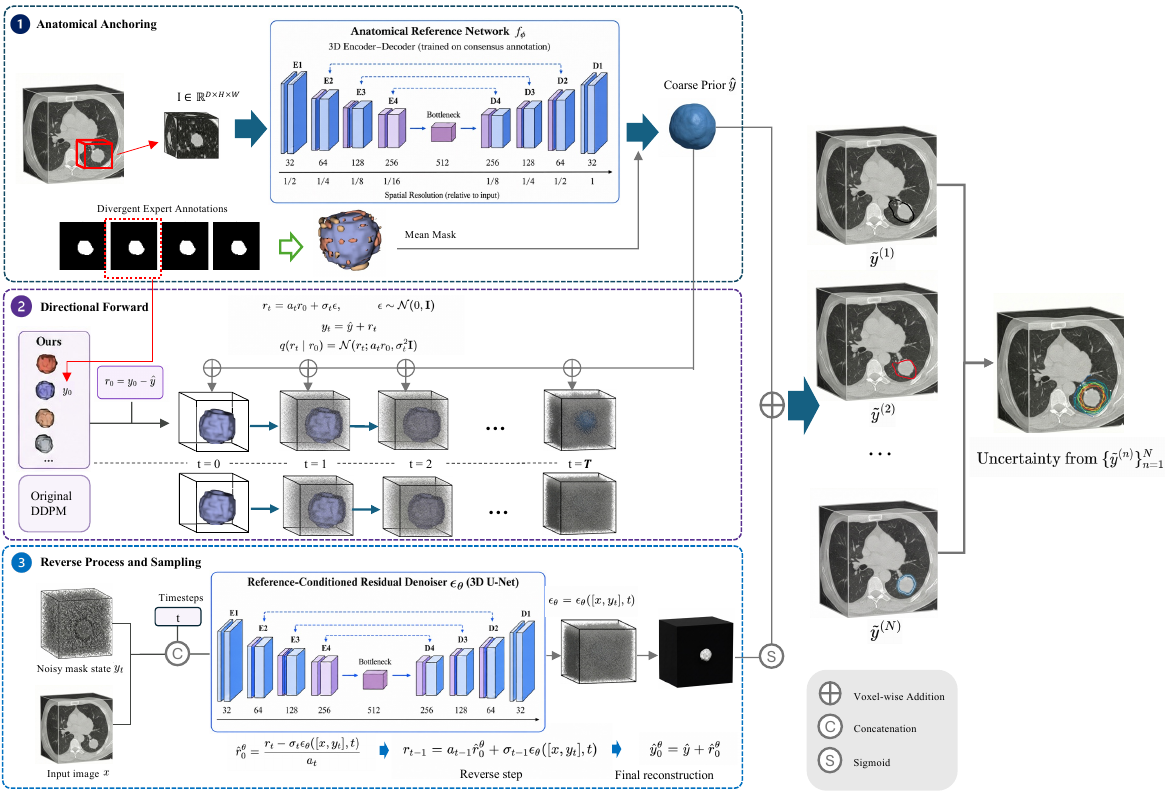}
\caption{Overview of the proposed Volumetric Directional Diffusion framework.}
\label{fig:method}
\end{figure*}

\subsection{Directional Diffusion in Residual Space}

Let $a_t$ and $\sigma_t$ denote the signal and noise coefficients of a fixed Gaussian corruption schedule, where $a_0\approx 1$, $\sigma_0\approx 0$, $a_T\approx 0$, and $\sigma_T\approx 1$. 
Following the DDPM-style denoising formulation~\cite{ho2020denoising}, a clean mask $y_0$ can be perturbed in closed form as
\begin{equation}
q(y_t \mid y_0)
=
\mathcal{N}\!\left(
y_t;
a_t y_0,
\sigma_t^2\mathbf{I}
\right).
\end{equation}
In our implementation, the coefficients are instantiated with a fixed linear schedule.

VDD applies this Gaussian corruption process to the residual field around the anatomical reference. 
Specifically, we define
\begin{equation}
r_0 = y_0-\hat{y}.
\end{equation}
Here, $r_0$ is not a segmentation mask itself, but a continuous residual field that represents the deviation between one expert annotation and the anatomical reference. 
The residual forward process is
\begin{equation}
q(r_t \mid r_0)
=
\mathcal{N}\!\left(
r_t;
a_t r_0,
\sigma_t^2\mathbf{I}
\right),
\end{equation}
or equivalently,
\begin{equation}
r_t
=
a_t r_0
+
\sigma_t\epsilon,
\quad
\epsilon\sim\mathcal{N}(0,\mathbf{I}).
\end{equation}

The corresponding noisy mask state is obtained by adding the anatomical reference back:
\begin{equation}
y_t = \hat{y}+r_t.
\end{equation}
Substituting $r_0=y_0-\hat{y}$ gives the VDD marginal in mask space:
\begin{equation}
q(y_t \mid y_0,\hat{y})
=
\mathcal{N}\!\left(
y_t;
a_t y_0+
(1-a_t)\hat{y},
\sigma_t^2\mathbf{I}
\right).
\end{equation}

Taking the conditional expectation of this VDD marginal gives
\begin{equation}
\mathbb{E}_{q(y_t\mid y_0,\hat{y})}[y_t]
=
a_t y_0+
(1-a_t)\hat{y}.
\end{equation}
As $a_t$ decreases along the forward process, the expected mask-space state moves from the expert annotation $y_0$ toward the anatomical reference $\hat{y}$. 
In particular, when $a_T \approx 0$, we have
\begin{equation}
\mathbb{E}_{q(y_T\mid y_0,\hat{y})}[y_T]\approx \hat{y}.
\end{equation}
Therefore, the terminal mask-space distribution is not centered at a case-independent zero-mean Gaussian state. 
Instead, it is centered around the case-specific anatomical reference, while the residual component approaches Gaussian noise.

This formulation preserves the Gaussian denoising principle of DDPM-style diffusion while shifting the corrupted variable from the full mask to the residual field around $\hat{y}$. 
Thus, VDD progressively corrupts expert-specific residuals without discarding the coarse anatomical support.
\subsection{Training Objective and Reverse Parameterization}

Using the residual forward process defined above, we sample one expert annotation $y_0$, a timestep $t$, and Gaussian noise $\epsilon$ to construct the noisy mask state $y_t$. 
The denoising network predicts the added residual noise from the concatenated image and noisy mask state, together with the timestep:
\begin{equation}
\epsilon_\theta = \epsilon_\theta([x,y_t],t).
\end{equation}
The anatomical reference $\hat{y}$ is not concatenated as an additional network input; instead, it conditions the denoising process implicitly through the construction of $y_t=\hat{y}+r_t$ and through the residual reconstruction steps.

We optimize a DDPM-style noise-prediction objective:
\begin{equation}
\mathcal{L}_{\mathrm{diff}}
=
\mathbb{E}_{y_0,t,\epsilon}
\left[
\left\|
\epsilon-\epsilon_\theta([x,y_t],t)
\right\|_2^2
\right].
\end{equation}

Given the predicted noise, we recover the predicted clean residual as
\begin{equation}
\hat{r}_0^\theta =
\frac{
r_t-\sigma_t\epsilon_\theta([x,y_t],t)
}{
a_t
}.
\end{equation}
The corresponding clean mask estimate is
\begin{equation}
\hat{y}_0^\theta = \hat{y}+\hat{r}_0^\theta.
\end{equation}

In practice, we optimize a hybrid objective by adding a time-weighted Focal Tversky~\cite{abraham2019novel} loss on the predicted clean mask. This auxiliary term provides direct shape supervision for the reconstructed foreground mask, complementing the noise-prediction loss that only supervises the residual denoising direction and is less sensitive to foreground sparsity and boundary quality:
\begin{equation}
\mathcal{L}
=
\mathcal{L}_{\mathrm{diff}}
+
\lambda w_t
\mathcal{L}_{\mathrm{FT}}\!\left(p_0^\theta, y_0\right),
\label{eq:hybrid_loss}
\end{equation}
where $p_0^\theta=\sigma(2\hat{y}_0^\theta)$ is the predicted foreground probability, $\sigma(\cdot)$ denotes the sigmoid function, and $\lambda$ is a dataset-dependent balancing coefficient. 
The timestep weight $w_t$ emphasizes the auxiliary segmentation loss at cleaner timesteps and reduces its influence at highly corrupted timesteps; in implementation, for the zero-based index $i\in\{0,\ldots,T-1\}$, we use $w_i=1-\frac{i+1}{T}$.

At inference time, the expert annotation $y_0$ is unavailable. 
We first obtain the anatomical reference $\hat{y}=f_{\phi}(x)$ and initialize the residual state as $r_T\sim\mathcal{N}(0,\mathbf{I})$, with $y_T=\hat{y}+r_T$. 
Using a DDIM-style reverse update in residual space, we compute
\begin{equation}
r_{t-1}
=
a_{t-1}\hat{r}_0^\theta
+
\sigma_{t-1}\epsilon_\theta([x,y_t],t),
\end{equation}
and recover the mask-space state as
\begin{equation}
y_{t-1}=\hat{y}+r_{t-1}.
\end{equation}
Repeating this process from the final timestep to $t=0$ produces one stochastic segmentation sample. 
Multiple samples are generated by independently sampling the initial residual noise $r_T$.
\subsection{Sampling and Uncertainty Estimation}

For each image, VDD generates $N$ stochastic segmentation samples
$\{\tilde{y}^{(n)}\}_{n=1}^{N}$ by repeating the reverse process with
independently sampled initial residual noise. Here, $\tilde{y}^{(n)}$ denotes
the final binary mask obtained from the $n$-th reverse sampling trajectory.
These samples are retained as multiple plausible segmentation hypotheses,
reflecting possible expert-specific boundary variations.

To summarize voxel-wise predictive variability, we compute the empirical
foreground probability
\begin{equation}
\hat{p}(v)=\frac{1}{N}\sum_{n=1}^{N}\tilde{y}^{(n)}(v),
\end{equation}
where $v$ indexes a voxel. The empirical foreground probability $\hat{p}(v)$ is used to summarize the predictive distribution and compute voxel-wise uncertainty maps. 
Dice-based sample-to-rater metrics are computed directly from the generated binary samples and expert annotations. 
When a consensus mask is needed for single-mask visualization or auxiliary boundary evaluation, we threshold $\hat{p}(v)$ at 0.5.

The voxel-wise uncertainty map is estimated as
\begin{equation}
U(v)=\hat{p}(v)\bigl(1-\hat{p}(v)\bigr).
\end{equation}

High values of $U(v)$ indicate locations where stochastic samples disagree,
which often correspond to ambiguous boundary regions.

\section{Experiments and Results}

\subsection{Datasets and Evaluation Metrics}
\begin{table*}[t]
\caption{Quantitative comparison across three multi-rater datasets. 
The best results are highlighted in \textbf{bold}, and the second-best are \underline{underlined}. 
GED is computed for single-output settings as a singleton prediction set; ``--'' indicates that CI/SNCC are not applicable without non-degenerate multi-sample predictive variance maps.}
\label{tab:comprehensive_results}
\centering
{\scriptsize
\setlength{\tabcolsep}{2.2pt}
\renewcommand{\arraystretch}{1.05}
\resizebox{\textwidth}{!}{
\begin{tabular}{lcc cccc cccc cccc}
\toprule
\multirow{2}{*}{\textbf{Method}} 
& \multirow{2}{*}{\textbf{Dim.}} 
& \multirow{2}{*}{\textbf{Type}} 
& \multicolumn{4}{c}{\textbf{LIDC-IDRI}} 
& \multicolumn{4}{c}{\textbf{KiTS21}} 
& \multicolumn{4}{c}{\textbf{ISBI 2015}} \\
\cmidrule(lr){4-7} \cmidrule(lr){8-11} \cmidrule(lr){12-15}
& & 
& \textbf{MaxDice$\uparrow$} & \textbf{GED$\downarrow$} & \textbf{CI$\uparrow$} & \textbf{SNCC$\uparrow$}
& \textbf{MaxDice$\uparrow$} & \textbf{GED$\downarrow$} & \textbf{CI$\uparrow$} & \textbf{SNCC$\uparrow$}
& \textbf{MaxDice$\uparrow$} & \textbf{GED$\downarrow$} & \textbf{CI$\uparrow$} & \textbf{SNCC$\uparrow$} \\
\midrule

nnUNet2    
& 2D & Det.  
& 0.6381 & 0.7198 & -- & --
& 0.7022 & 0.7623 & -- & --
& 0.6775 & 0.7145 & -- & -- \\

DiffOSeg   
& 2D & Diff. 
& 0.5969 & 0.4803 & 0.2406 & \underline{0.3444}
& 0.5804 & \underline{0.6511} & 0.1542 & 0.1542
& 0.5551 & 0.5622 & 0.1785 & 0.1980 \\

CCDM       
& 2D & Diff. 
& 0.2531 & 0.6992 & 0.0000 & 0.1992
& 0.5632 & 0.6842 & 0.1235 & \underline{0.1844}
& 0.5378 & \underline{0.5217} & \underline{0.1792} & \underline{0.2931} \\

\midrule

nnUNet2    
& 3D & Det.  
& \underline{0.7622} & \underline{0.4744} & -- & --
& \underline{0.7401} & 0.6526 & -- & --
& \underline{0.6914} & 0.6787 & -- & -- \\

Prob U-Net 
& 3D & Prob. 
& 0.7297 & 0.5467 & \underline{0.2906} & 0.1078
& 0.5908 & 0.9634 & \underline{0.2120} & 0.0287
& \textbf{0.7667} & 0.5595 & 0.0871 & 0.0665 \\

VDD (Ours) 
& 3D & Diff. 
& \textbf{0.8031} & \textbf{0.2081} & \textbf{0.6360} & \textbf{0.5322}
& \textbf{0.9145} & \textbf{0.1956} & \textbf{0.7233} & \textbf{0.2509}
& 0.6345 & \textbf{0.4798} & \textbf{0.2605} & \textbf{0.2957} \\

\bottomrule
\end{tabular}
}}
\end{table*}

We evaluated VDD on three multi-rater datasets: LIDC-IDRI~\cite{armato2011lung}, KiTS21~\cite{heller2023kits21}, and ISBI 2015~\cite{carass2017longitudinal}. 
All datasets were partitioned at the patient level using an 8:1:1 train/validation/test split to prevent patient-level data leakage. 
We assess VDD from two complementary perspectives: sample-to-rater agreement and uncertainty-disagreement alignment. 
For Dice-based evaluation, all stochastic methods are evaluated with the same number of samples, $N=16$, unless otherwise specified. 
Given $N$ predicted samples and $K$ expert annotations for the same case, MeanDice averages Dice over all $N \times K$ prediction--annotation pairs, while MaxDice first matches each expert annotation to its best generated sample and then averages over the $K$ annotations. 
Thus, MeanDice reflects average sample-to-rater agreement, whereas MaxDice measures sample-set coverage of plausible rater-specific delineations. 
We additionally report 95\% Hausdorff Distance (HD95) for volumetric boundary consistency~\cite{maierhein2024metrics}.

\subsection{Baseline Coverage}

We compare VDD against representative baselines along the main implementation axes in ambiguous medical image segmentation. 
First, nnUNet2 is included in both 2D and 3D forms as a strong deterministic reference. 
This comparison separates the effect of volumetric modeling from stochastic uncertainty estimation: deterministic models typically provide strong single-mask accuracy, but they do not directly generate multiple plausible annotations. 
Second, Probabilistic U-Net represents latent-variable probabilistic segmentation, where sample diversity is introduced through a learned latent space rather than an iterative diffusion process. 
Third, CCDM and DiffOSeg represent diffusion-based stochastic segmentation baselines. 
They are included to evaluate whether diffusion-based sampling can capture inter-rater variability and how much volumetric consistency is retained when uncertainty is generated slice-wise or through multi-expert diffusion modeling. This baseline set therefore covers deterministic segmentation, latent probabilistic segmentation, and diffusion-based stochastic segmentation, as well as both 2D and 3D implementations. 

\subsection{Comparison with Deterministic and Generative Baselines}
% Preamble:
% \usepackage{booktabs}
% \usepackage{graphicx}

On LIDC-IDRI, VDD achieves the strongest performance across all metrics reported in Table~\ref{tab:comprehensive_results}. 
It obtains the highest MaxDice of 0.8031, compared with 0.7622 from 3D nnUNet2 and 0.7297 from Prob U-Net, indicating stronger sample-to-rater coverage of plausible expert delineations. 
VDD also substantially reduces GED to 0.2081, compared with 0.4744 from 3D nnUNet2, 0.5467 from Prob U-Net, and 0.4803 from DiffOSeg. 
In addition, VDD achieves the highest CI and SNCC, suggesting that its predictive uncertainty is better aligned with regions of expert disagreement.
\begin{table}[!htbp]
\caption{Representative 2D-to-3D evaluation on LIDC-IDRI.}
\label{tab:lidc_2d3d}
\centering
{
\setlength{\tabcolsep}{1.3pt}
\begin{tabular}{llcccccc}
\toprule
\textbf{Method} & \textbf{Eval.} 
& \textbf{MeanDice$\uparrow$} 
& \textbf{MaxDice$\uparrow$} 
& \textbf{HD95$\downarrow$} 
& \textbf{GED$\downarrow$} 
& \textbf{CI$\uparrow$} 
& \textbf{SNCC$\uparrow$} \\
\midrule
\multirow{2}{*}{CCDM}
& 2D & 0.5077 & 0.7462 & 6.0780  & {0.3357} & 0.1182 & {0.4810} \\
& 3D & 0.2185 & 0.2531 & 27.6041 & 0.6992 & 0.0000 & 0.1992 \\
\midrule
\multirow{2}{*}{DiffOSeg}
& 2D & \underline{0.6185} & \underline{0.7532} & \underline{3.9622} & \underline{0.3108} & \underline{0.5280} & \underline{0.5103} \\
& 3D & 0.5060 & 0.5969 & 16.0721 & 0.4803 & 0.2406 & 0.3444 \\
\midrule
VDD & 3D & \textbf{0.7142} & \textbf{0.8031} & \textbf{2.1712} & \textbf{0.2081} & \textbf{0.6360} & \textbf{0.5322} \\
\bottomrule
\end{tabular}}
\end{table}
On KiTS21, VDD also achieves the best performance across all metrics reported in Table~\ref{tab:comprehensive_results}. 
It obtains the highest MaxDice of 0.9145, compared with 0.7401 from 3D nnUNet2 and 0.7022 from 2D nnUNet2. 
For distributional uncertainty modeling, VDD reduces GED to 0.1956, compared with 0.6511 from DiffOSeg, 0.6842 from CCDM, and 0.9634 from Prob U-Net. 
It also achieves the highest CI of 0.7233 and the highest SNCC of 0.2509, indicating stronger agreement between the generated sample distribution and the observed inter-rater variability.

On ISBI 2015, Prob U-Net achieves the highest MaxDice of 0.7667, while VDD obtains a lower MaxDice of 0.6345. 
However, VDD achieves the strongest uncertainty-disagreement alignment among the evaluated methods, with the lowest GED of 0.4798, the highest CI of 0.2605, and the highest SNCC of 0.2957. 
This result suggests that VDD is not optimized solely for maximizing best-sample overlap, but for generating plausible segmentation samples whose distribution better reflects expert disagreement. 
Table~\ref{tab:comprehensive_results} summarizes the quantitative comparison across three multi-rater datasets. 
Overall, VDD consistently improves distributional uncertainty alignment, while achieving strong sample-to-rater coverage on LIDC-IDRI and KiTS21 and remaining competitive on ISBI 2015.
Table~\ref{tab:lidc_2d3d} further examines whether the gap between VDD and 2D diffusion baselines is only due to the use of 3D evaluation.

Since CCDM and DiffOSeg operate slice-wise, stacking their outputs into 3D volumes may introduce additional penalties from through-plane inconsistency. 
We therefore also evaluate them in their native 2D setting, which provides a favorable comparison for slice-wise methods. 
Even under this setting, VDD evaluated directly in 3D outperforms both CCDM and DiffOSeg on LIDC-IDRI across MeanDice, MaxDice, HD95, GED, CI, and SNCC. 
This suggests that the advantage of VDD is not merely an artifact of 3D reconstruction, but comes from modeling stochastic boundary variation within a coherent volumetric space.

% Table 1: main quantitative comparison
% 必须包含 existing baselines + BerDiff/HiDiff/AR-Seg placeholders + SDEdit-style placeholder

\subsection{Ablation and Rater Stability}
Table~\ref{tab:ablation} evaluates the effect of anatomical anchoring and stochastic sampling on LIDC-IDRI. 
Removing the anatomical reference substantially degrades both sample-to-rater agreement and uncertainty alignment: at $N=16$, MaxDice decreases from 0.8031 to 0.6670, HD95 increases from 2.1712 to 13.1035, and GED increases from 0.2081 to 0.5809. 
This indicates that the anatomical reference provides a necessary volumetric support for residual diffusion. 
Importantly, increasing the number of samples improves MaxDice, GED, CI, and SNCC, showing that VDD does not simply copy the prior but uses stochastic residual sampling to cover plausible rater-specific boundary variations.
\begin{table}[!htbp]
\caption{Ablation study evaluating anatomical anchoring and the number of stochastic samples on LIDC-IDRI.}
\label{tab:ablation}
\centering
{\scriptsize
\setlength{\tabcolsep}{2.0pt}
\begin{tabular}{lccccccc}
\toprule
\textbf{Method} & \textbf{$N$} & \textbf{MeanDice$\uparrow$} & \textbf{MaxDice$\uparrow$} & \textbf{HD95$\downarrow$} & \textbf{GED$\downarrow$} & \textbf{CI$\uparrow$} & \textbf{SNCC$\uparrow$} \\
\midrule
VDD w/o anchor & 1  & 0.4916 & 0.4916 & 12.6894 & 1.0010 & --     & --     \\
VDD w/o anchor & 4  & 0.4830 & 0.6087 & 13.1829 & 0.6531 & 0.3743 & 0.2831 \\
VDD w/o anchor & 16 & 0.4867 & 0.6670 & 13.1035 & 0.5809 & 0.4469 & 0.3393 \\
\midrule
VDD & 1  & \underline{0.7177} & 0.7177 & \underline{2.1668} & 0.5762 & --     & --     \\
VDD & 4  & \textbf{0.7192} & \underline{0.7735} & \textbf{2.0424} & \underline{0.2748} & \underline{0.5408} & \underline{0.4412} \\
VDD & 16 & 0.7142 & \textbf{0.8031} & 2.1712 & \textbf{0.2081} & \textbf{0.6360} & \textbf{0.5322} \\
\bottomrule
\end{tabular}}
\end{table}
\begin{figure*}[t]
\centering
\includegraphics[width=\textwidth]{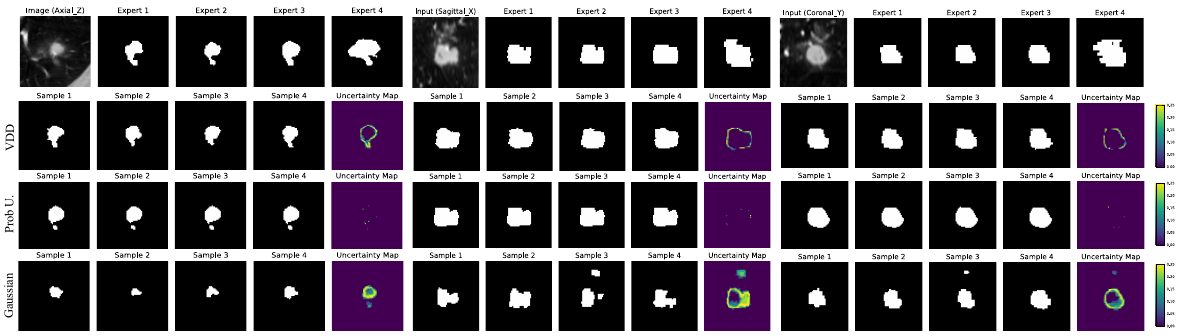}
\caption{Visual comparison of uncertainty maps among VDD, Prob U-Net, and VDD (Gaussian).}
\label{fig:uncertain}
\end{figure*}
Table~\ref{tab:rater} further examines whether VDD remains stable when the annotation set becomes more diverse. 
As the number of raters increases from $K=2$ to $K=4$, MeanDice changes only slightly from 0.7203 to 0.7142, while MaxDice remains nearly unchanged around 0.803. 
This suggests that adding more expert opinions expands the ambiguity of the target distribution, but does not collapse the sample-set coverage of VDD.
\begin{table}[!htbp]
\caption{Stability analysis of VDD on LIDC-IDRI under different numbers of expert annotations.}
\label{tab:rater}
\centering
{
\setlength{\tabcolsep}{2.4pt}
\begin{tabular}{lcccccc}
\toprule
\textbf{$K$} & \textbf{MeanDice$\uparrow$} & \textbf{MaxDice$\uparrow$} & \textbf{GED$\downarrow$} & \textbf{HD95$\downarrow$} & \textbf{CI$\uparrow$} & \textbf{SNCC$\uparrow$} \\
\midrule
$K=2$ & 0.7203 & 0.8030 & 0.2034 & 1.8689 & 0.6287 & 0.5364 \\
$K=3$ & 0.7154 & 0.8037 & 0.2018 & 2.0494 & 0.6281 & 0.5359 \\
$K=4$ & 0.7142 & 0.8031 & 0.2081 & 2.1712 & 0.6360 & 0.5322 \\
\bottomrule
\end{tabular}}
\end{table}

The uncertainty metrics show a similar trend. 
GED remains close to 0.20 across all settings, and CI/SNCC also stay stable, with CI slightly increasing to 0.6360 at $K=4$ and SNCC changing only mildly from 0.5364 to 0.5322. 
These results suggest that VDD can absorb increased rater diversity while maintaining consistent uncertainty-disagreement alignment, which is central to ambiguous multi-rater segmentation.

% Table 2: anchoring, N, prior sensitivity, rater-focused metrics
% 这里放 Reviewer #1 和 Reviewer #2 关心的实验

\subsection{Visual Analysis of Uncertainty and Residual Exploration}
% Fig. 2 / Fig. 3

\begin{figure*}[t]
\centering
\includegraphics[width=0.8\textwidth]{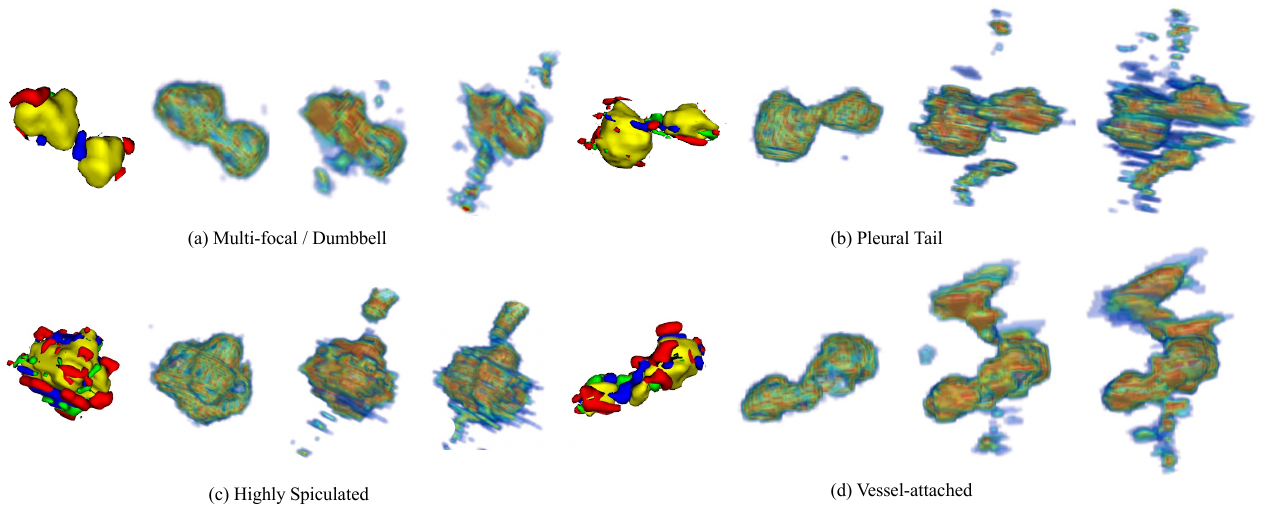}
\caption{3D visual comparison of uncertainty estimation on representative ambiguous LIDC-IDRI nodules, including (a) multi-focal/dumbbell, (b) pleural-tail, (c) highly spiculated, and (d) vessel-attached cases. Each case compares multi-rater ground-truth annotations with uncertainty maps from VDD, CCDM, and DiffOSeg.}
\label{fig:3d_vis}
\end{figure*}

\begin{figure}[t]
\centering
\includegraphics[width=\columnwidth]{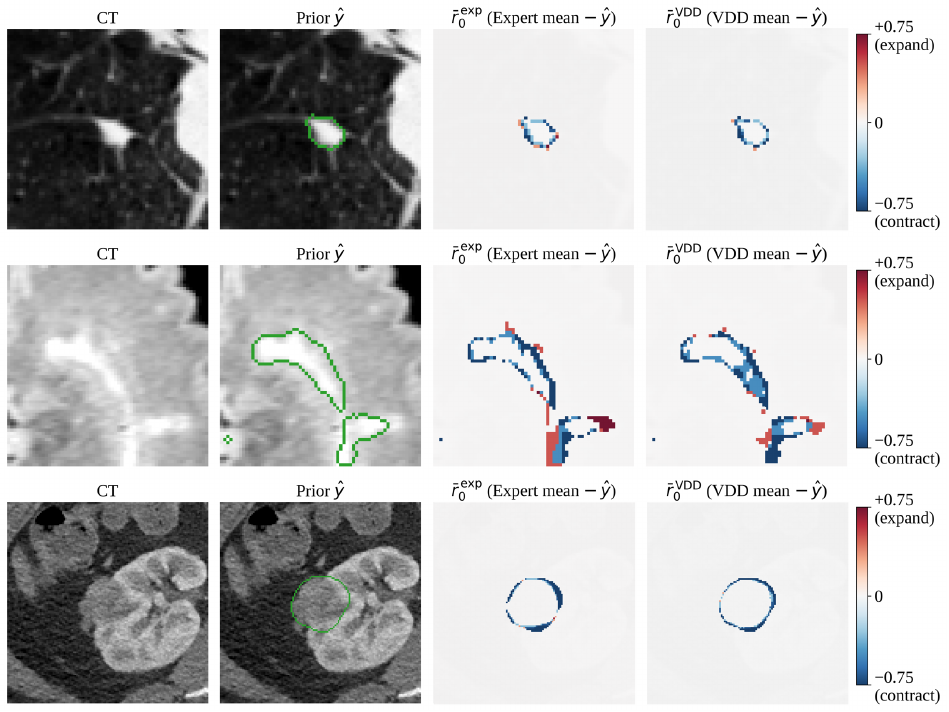}
\caption{Visualization of residual boundary exploration. For each case, the anatomical prior provides coarse object support, while the residual field concentrates around uncertain boundary regions.}
\label{fig:residual}
\end{figure}

Figs.~\ref{fig:uncertain}--\ref{fig:residual} provide a progressive visual analysis of VDD, from uncertainty behavior to volumetric consistency and residual boundary correction. 
Fig.~\ref{fig:uncertain} first examines whether stochastic models place uncertainty in clinically meaningful regions. 
We do not include nnUNet2 in this visualization because deterministic models only produce a single mask and therefore cannot provide non-degenerate predictive uncertainty maps. 
Instead, we compare VDD with two representative stochastic extremes across axial, sagittal, and coronal views. 
Although Prob U-Net is designed for ambiguous segmentation, its uncertainty maps are mostly inactive, with only sparse responses near a few boundary locations. 
This suggests that its sampled masks tend to collapse to similar predictions and may under-explore plausible expert disagreement. 
In contrast, the Gaussian variant of VDD removes the anatomical prior and starts from unstructured Gaussian noise, resembling a conventional full-mask DDPM setting. 
While this prior-free variant produces stronger diversity, its uncertainty also spreads into the interior consensus region where experts largely agree, indicating over-exploration of anatomically stable regions. 
VDD provides a more balanced behavior: uncertainty is concentrated around equivocal boundaries, while the common lesion support remains stable.

Fig.~\ref{fig:3d_vis} further examines whether diffusion-based uncertainty remains coherent in 3D. 
The leftmost colored surfaces show different expert annotations, while the following 3D overlays are obtained by first aggregating 16 generated masks into probability/uncertainty volumes and then aligning them with the original image space for visualization in ITK-SNAP. 
Thus, the overlays represent volumetric predictive uncertainty rather than individual sampled segmentations. 
Blue/cyan indicates low uncertainty, green/yellow indicates moderate uncertainty, and orange/red indicates high uncertainty. 
For CCDM and DiffOSeg, the uncertainty maps often show strong variation, but the corresponding 3D structures are fragmented, jagged, or spatially redundant. 
This is consistent with their slice-wise generation process: adjacent slices are sampled independently, so uncertainty may become discontinuous along the through-plane direction. 
Such instability is especially visible in nodules with thin bridges, pleural tails, vessel attachments, or highly spiculated boundaries, where fragmented or redundant uncertain regions may obscure the distinction between lesion boundary and surrounding anatomy. 
In contrast, VDD preserves coherent volumetric support while assigning uncertainty mainly to boundary regions where expert annotations diverge. 
These visual patterns support the quantitative trends in Tables~\ref{tab:comprehensive_results} and~\ref{tab:lidc_2d3d}, suggesting that VDD improves uncertainty-disagreement alignment without sacrificing volumetric consistency.
Fig.~\ref{fig:residual} further visualizes how VDD performs residual boundary correction around the anatomical prior on three representative cases from different datasets. 
The green contour denotes the anatomical prior, which provides coarse object support but is not treated as a perfect pseudo-label: it may over-cover normal tissue in some regions while under-covering fine ambiguous structures in others. 
Here, the ``mean'' masks are soft probability maps rather than binary masks. 
The expert mean is obtained by averaging multiple expert annotations, while the VDD mean is the empirical foreground probability averaged over $N=16$ stochastic samples. 
We subtract the same anatomical prior from both soft means to obtain residual maps: positive values indicate expansion beyond the prior, whereas negative values indicate contraction relative to the prior.

As shown in Fig.~\ref{fig:residual}, the expert residual reveals where human annotations tend to expand or suppress the coarse prior. 
The VDD residual follows a similar spatial pattern, expanding regions that are also expanded by the expert mean and suppressing regions that are contracted by the expert mean. 
This residual-level agreement shows that VDD does not simply copy the anatomical prior or freely explore the full mask space. 
Instead, it learns expert-like boundary corrections around the prior, preserving clinically meaningful disagreement while maintaining anatomical support.

\subsection{Clinical Applicability: Inference Efficiency}
% Table 3 or Table 4 depending on numbering

Diffusion-based stochastic segmentation is inherently more computationally demanding than deterministic segmentation because multiple iterative denoising steps are required for each sampled mask. 
Therefore, Table~\ref{tab:efficiency} compares VDD with representative diffusion-based stochastic segmentation methods under the same efficiency category, rather than with deterministic single-pass models. 
All methods are evaluated on a single H100 GPU with I/O overhead excluded. 
Under their native inference settings, VDD requires 0.1506 seconds and 0.39 GB memory for one inference pass, compared with 1.2404 seconds and 1.12 GB for CCDM and 1.5199 seconds and 0.61 GB for DiffOSeg. 
When generating 16 stochastic samples for uncertainty estimation, VDD requires 2.41 seconds, whereas CCDM and DiffOSeg require 19.85 and 24.32 seconds, respectively.

This efficiency gain suggests that anatomical anchoring reduces the effective denoising burden by constraining stochastic generation to residual boundary exploration around a coarse anatomical reference. 
Instead of reconstructing a complete segmentation mask from a broad noisy state, VDD only needs to sample plausible residual deviations around the anatomical support. 
From a clinical workflow perspective, the reduced sampling cost makes repeated 3D sampling more feasible for generating voxel-wise uncertainty maps, which may support downstream review in tasks such as radiotherapy planning, surgical margin assessment, or risk-aware lesion delineation.

\begin{table}[!htbp]
\caption{Inference efficiency comparison among diffusion-based stochastic segmentation methods.}
\label{tab:efficiency}
\centering
{\footnotesize
\setlength{\tabcolsep}{2.5pt}
\begin{tabular}{lccccc}
\toprule
\textbf{Method} & \textbf{Params} & \textbf{Steps}\footnotemark & \textbf{Time/pass} & \textbf{Mem.} & \textbf{Time$\times$16} \\
 & \textbf{(M)} & \textbf{$\downarrow$} & \textbf{(s)$\downarrow$} & \textbf{(GB)$\downarrow$} & \textbf{(s)$\downarrow$} \\
\midrule
CCDM       & 14.4 & 250 & $1.2404{\pm}0.0027$ & 1.12 & 19.85 \\
DiffOSeg   & 25.7 & 250 & $1.5199{\pm}0.0123$ & 0.61 & 24.32 \\
VDD (Ours) & 20.1 & \textbf{50} & $\mathbf{0.1506{\pm}0.0004}$ & \textbf{0.39} & \textbf{2.41} \\
\bottomrule
\end{tabular}}
\end{table}
\footnotetext{CCDM and DiffOSeg are evaluated using their official 250-step diffusion configuration.}
\section{Conclusion}
We presented Volumetric Directional Diffusion (VDD), a prior-anchored diffusion framework for uncertainty-aware ambiguous 3D medical image segmentation. 
By modeling expert-specific variations as residual deviations around a coarse anatomical reference, VDD shifts stochastic generation from full-mask synthesis to boundary-focused residual exploration. 
Experiments on three multi-rater datasets show that VDD improves uncertainty alignment, while maintaining competitive volumetric segmentation accuracy. 
The ablation study further suggests that anatomical anchoring helps stabilize stochastic sampling and preserve structural consistency. Future work will further study robustness to imperfect anatomical references and broader adaptation of prior-anchored residual diffusion to other multi-rater volumetric segmentation settings.

\bibliographystyle{IEEEtran}
\bibliography{mybibliography}

\end{document}